\documentclass{article}
\ifdefined\pdfminorversion
\fi
\usepackage{preprint_layout,times}

\usepackage{amsmath,amsfonts,bm}

\def\eqref#1{equation~\ref{#1}}

\def\1{\bm{1}}

\DeclareMathAlphabet{\mathsfit}{\encodingdefault}{\sfdefault}{m}{sl}
\SetMathAlphabet{\mathsfit}{bold}{\encodingdefault}{\sfdefault}{bx}{n}

\usepackage[hidelinks]{hyperref}
\usepackage{url}
\usepackage{booktabs}
\usepackage{graphicx}

\title{What You Show Is Not What You Get: Rethinking Demonstration Following in Visual In-Context Segmentation}

\author{
Xiao Hu \qquad Ilia Sucholutsky\\
Purdue University\\
\texttt{\{hu953,isucholu\}@purdue.edu}
}

\hypersetup{
  pdftitle={What You Show Is Not What You Get: Rethinking Demonstration Following in Visual In-Context Segmentation},
  pdfauthor={Xiao Hu, Ilia Sucholutsky}
}

\begin{document}

\maketitle

\begin{abstract}
Visual in-context segmentation uses visual demonstrations to guide predictions for a new image. In SegGPT, we identify background color mapping deviation, where non-target regions sharing a background label receive different output colors, and part-to-parent expansion, where predictions extend beyond the requested part into its parent object. These deviations can occur even when the relevant regions remain distinguishable in internal features. Controlled changes to demonstration colors and requested targets support an account in which demonstrations guide role and color assignments over familiar semantic regions, influenced by learned semantic and color preferences. Building on these findings, we introduce Semantic Contrast Prompt Selection (SCPS). Its semantic contrast encoding assigns separate colors to the requested part and the rest of its object while keeping the remaining scene black. SCPS combines existing retrieval scores with quality estimates from labeled development images to select demonstrations and their binary or semantic contrast encodings. Averaged across seven retrieval methods on three benchmarks, SCPS improves target IoU by 0.054 and reduces leakage into the rest of the object by 22.7\% relative to the original binary selectors. Our findings offer a new perspective on understanding and improving demonstration following in visual in-context segmentation.
\end{abstract}

\section{Introduction}
In-context learning (ICL) enables models to perform tasks from demonstrations without parameter updates~\citep{min2022rethinking}. Visual in-context learning (VICL) extends this paradigm to vision through paired input and output images that guide predictions for a query image~\citep{NEURIPS2022_9f09f316,Wang_2023_CVPR}. Research on language ICL distinguishes task recognition, which elicits familiar task knowledge, from task learning, which infers mappings unseen during pretraining~\citep{pan2023context}. In VICL models supporting heterogeneous tasks, activations associated with specific tasks can guide predictions without the original demonstrations~\citep{VTC}, while unfamiliar demonstrated encodings can elicit familiar task outputs or encodings used during training~\citep{khatri2026modelsizeprobinggaps}. These findings suggest that visual demonstrations can elicit learned behavior without ensuring adherence to the demonstrated mapping. In visual in-context segmentation, training strategies such as randomization of label colors encourage reliance on demonstrations within a single task family~\citep{Wang_2023_ICCV}. Whether such training leads to faithful demonstration following remains an open question.

\begin{figure}[t]
\centering
\includegraphics[width=0.75\linewidth]{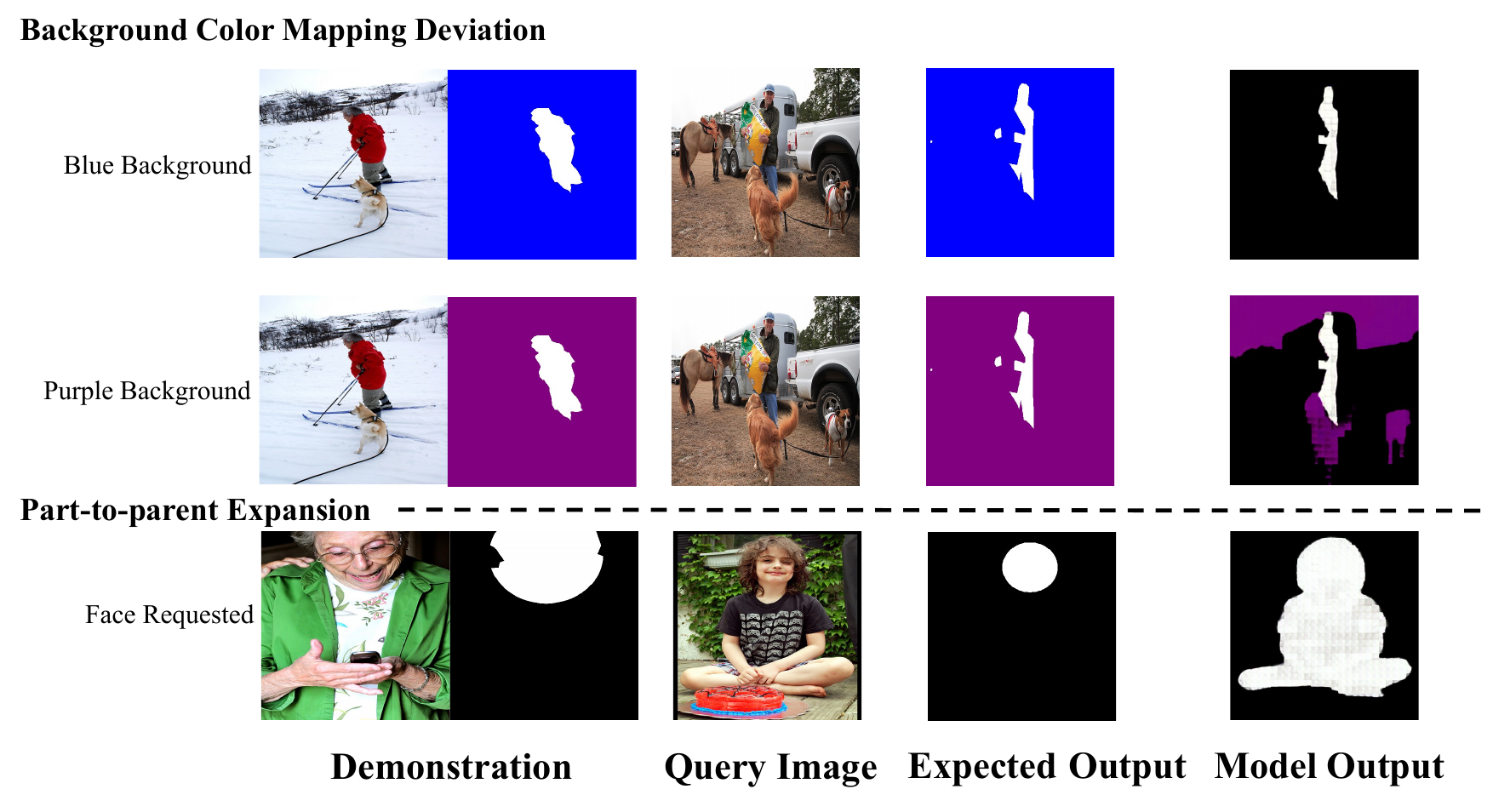}
\caption{\textbf{Deviations from demonstration following.} Expected outputs apply the demonstrated colors to the annotated query regions. Top: non-target regions sharing the same background label receive different output colors. Bottom: a demonstration marking only the face yields a prediction extending to the whole person.}
\label{fig1}
\end{figure}

A segmentation demonstration specifies both target regions and output encodings. We study adherence to these specifications through RGB predictions and identify two deviations, illustrated in Figure~\ref{fig1}. First, although requested targets generally receive the demonstrated foreground color, non-target regions sharing the same background label can receive different output colors. Some receive the demonstrated non-black background color, while others remain black. We refer to this selective behavior as \emph{background color mapping deviation}. Second, a prediction can extend beyond the demonstrated part into its enclosing object, which we refer to as the parent object. We call this \emph{part-to-parent expansion}. This occurs even though the demonstration mask explicitly labels the rest of the parent object as background.

To understand background color mapping deviation, we vary demonstration colors and requested targets while examining predictions and internal representations. Non-target regions respond differently depending on their semantic identity, demonstrated role, and color, yet remain distinguishable from the remaining scene in late features. These findings support a mechanistic account in which the model recognizes familiar semantic regions and uses demonstrations to assign their roles and colors. Learned semantic and color preferences influence how these assignments appear in the output.

Our findings on selective background coloring also help explain part-to-parent expansion. Linear probes trained separately for each task on disjoint queries distinguish the requested part from the \emph{parent-only region}, defined as the parent object excluding that part, in most analyzed expansion cases, although both receive the foreground color. A shared probe further shows that late features retain information about which part the demonstration requests. In face segmentation, changing the demonstrated background color selectively recolors much of the body while leaving the remaining scene predominantly black. Together, these findings connect part-to-parent expansion to how demonstration cues affect the output colors of internally distinguishable regions.

The generally reliable coloring of requested targets, together with the response of parent-only regions to background color changes, motivates explicitly marking the parent-only region with its own color. We call this a \emph{semantic contrast encoding}, which assigns distinct colors to the target, parent-only region, and remaining scene. We keep the remaining scene black, consistent with its tendency to remain black in our experiments, and separately color the parent-only region to cue its exclusion from the target. Although this encoding reduces parent leakage, applying it uniformly can impair target segmentation. We therefore introduce Semantic Contrast Prompt Selection (SCPS), which combines existing selection scores with target quality estimates computed on development data to jointly select a demonstration and its binary or semantic contrast encoding. 

Aggregated across three part segmentation benchmarks~\citep{Gong_2017_CVPR,he2022partimagenet,wei2023ovparts}, SCPS improves target intersection-over-union (IoU) and Strict Target Success (STS) while reducing Parent Leakage (PL) relative to the original selectors for all seven evaluated retrieval methods. PL measures the fraction of the parent-only region incorrectly predicted as foreground, while STS requires accurate target segmentation and low leakage. Averaged across the seven methods, SCPS increases target IoU by $0.054$, reduces PL by $22.7\%$ relative to the corresponding baselines, and increases STS by $6.37$ percentage points.

Our contributions are as follows:
\begin{itemize}
\item We identify background color mapping deviation and part-to-parent expansion in visual in-context segmentation. We propose a mechanistic account linking familiar semantic regions, role and color assignments guided by demonstrations, and learned semantic and color preferences.

\item We show that semantic regions and the target roles specified by demonstrations can remain readable in late features even when predictions violate the demonstration. Controlled changes to colors and requested targets, together with diagnostic feature interventions, connect these internal distinctions to output behavior.

\item We introduce Semantic Contrast Prompt Selection (SCPS), which combines semantic contrast encoding with target quality calibration to jointly select demonstrations and their binary or semantic contrast encodings. Aggregated across three benchmarks, SCPS improves target IoU and STS while reducing PL for each of seven retrieval methods.
\end{itemize}

\section{Related Work}
\textbf{Visual In-Context Learning.}
VICL encompasses frameworks supporting heterogeneous tasks through image inpainting~\citep{NEURIPS2022_9f09f316}, unified image prediction~\citep{Wang_2023_CVPR}, autoregressive modeling~\citep{Bai_2024_CVPR,pmlr-v235-hao24b}, and fusion of multiple prompts~\citep{liao2026enhancingvisualincontextlearning}, alongside approaches specializing in low-level vision~\citep{pmlr-v235-liu24k,chen2024learning,Rajagopalan2024AWRaCLeAI} or segmentation. Within segmentation, SegGPT randomizes label colors to encourage reliance on demonstrations~\citep{Wang_2023_ICCV}, while SINE predicts multiple masks for different segmentation tasks to address task ambiguity~\citep{NEURIPS2024_2cc0b084}.\footnote{SINE reports part segmentation results for a variant trained with additional part annotations, but the corresponding checkpoint was not available in its official repository at the time of our experiments.} We focus on SegGPT, a representative model trained on diverse segmentation data, including semantic, instance, and part segmentation~\citep{Wang_2023_ICCV}, and integrated into automatic annotation tools such as Autodistill~\citep{autodistill2023seggpt}. Its broad training coverage and randomized label colors make it a relevant testbed for studying demonstration following, while its RGB outputs enable joint analysis of region selection and color mapping.

\textbf{Demonstration Selection.}
UnsupPR~\citep{zhang2023goodexamples} retrieves demonstrations by similarity in frozen visual features. Our evaluation also includes retrieval based on CLIP and DINOv2 features~\citep{radford2021clip,oquab2023dinov2}. Learned retrievers include SupPR~\citep{zhang2023goodexamples}, LaPR~\citep{luo2026lapr}, Partial2Global, and RH-Partial2Global~\citep{xu2024partial2global,wu2025rhpartial2global}. SCPS augments existing demonstration selectors by jointly selecting demonstrations and their binary or semantic contrast encodings, combining existing selection scores with target quality estimates computed on development data.

\section{Problem Setup}
\subsection{Segmentation Demonstrations and Color Encodings}
\label{sec:demonstrations_encodings}

A demonstration pairs an image $x_d$ with an RGB segmentation map $y_d$. Given a query image $x_q$, a frozen model predicts $\hat{y}_q = F_{\theta}(x_q; x_d, y_d)$. For the two-color demonstrations used in our analysis, the requested target receives foreground color $c_f$, and all other pixels receive background color $c_b$. The query prediction is expected to follow the same mapping. The standard binary demonstrations use white foreground and black background.
Regions sharing the background label can include non-target objects and the remaining scene. We examine them separately. For a requested part $T$ within its parent object $P$, the parent-only region is $P \setminus T$, and the remaining scene lies outside $P$.

For these color comparisons, we assign each predicted pixel to the nearest demonstrated foreground or background color using Euclidean distance in RGB space. When the background color is non-black, we also include black as a candidate color. Color adherence is the fraction of pixels in an annotated query region assigned to its demonstrated color.

\subsection{Controlled Experiments and Feature Probes}
\label{sec:controlled_analysis}

We keep the demonstration image, target mask, and query fixed while changing the foreground or background color in the demonstration. We also compare requesting different objects while preserving each object's assigned color. For example, requesting the person assigns gray to the person and magenta to all other pixels; requesting the dog assigns magenta to the dog and gray to all other pixels.

We extract query features from SegGPT's final transformer block during the same forward pass as the RGB prediction. For the background-color experiment, we also extract features from the first transformer block to compare early and late representations. Linear probes are classifiers trained to distinguish annotated regions using these features. We train them on one set of queries and evaluate them on held-out queries using the area under the receiver operating characteristic curve (AUROC). They distinguish non-target semantic regions from the remaining scene, or requested parts from parent-only regions. In the color comparisons, we train a separate probe for each demonstration color setting while keeping the annotated region definitions unchanged. This lets us examine whether those regions remain distinguishable internally as their output colors change. Appendix~\ref{app:analysis-details} provides the detailed analysis protocols and statistical procedures.

\section{Understanding Deviations from Demonstration Following}
We first examine how demonstration colors and requested targets affect background coloring. We then ask whether the affected regions remain distinguishable internally and connect these findings to part-to-parent expansion.

\subsection{Background Color Mapping Deviation}
\label{sec:background_deviation}

Figure~\ref{fig1} shows that regions sharing the background label can receive different output colors. To examine this behavior across queries, we fix one demonstration marking the person as the target and evaluate 15 held-out COCO query images containing a person and a dog \citep{lin2014microsoft} under eight background colors, keeping the foreground white. As shown in Figure~\ref{fig:background_features}, purple produces the highest mean adherence to the demonstrated background color among dog pixels, at 6.3\%. Some pixels in the remaining scene, which also contains other semantic regions, receive the demonstrated background color. Most pixels in both regions remain black on average. Thus, the dog and the remaining scene respond differently to the demonstrated background color despite sharing the same background label.

We next examine how changing the requested object affects coloring while keeping each object's assigned color unchanged. In one fixed demonstration and query pair containing a person and a dog, we assign gray to the person and magenta to the dog. The dog receives magenta on $0.0\%$ of its pixels as a non-target but $90.0\%$ as the target, while the person remains predominantly gray in both cases. We extend this comparison to 30 queries, three demonstrations, and four fixed color pairs. After averaging across demonstrations and color pairs within each query, the mean increase in adherence is $21.1$ percentage points for the dog, whereas the person shows a mean decrease of $2.5$ percentage points. Appendix~\ref{app:requested-object} provides the protocol and confidence intervals.

Thus, requesting a region can change whether it receives its assigned color, but the direction and magnitude differ between the person and dog. These observations suggest that background coloring depends jointly on semantic identity, demonstrated role, and color encoding.

\subsection{Semantic Distinctions Behind Color Changes}
\label{sec:semantic_distinctions}

\begin{figure}[t]
    \centering
    \includegraphics[width=0.7\linewidth]{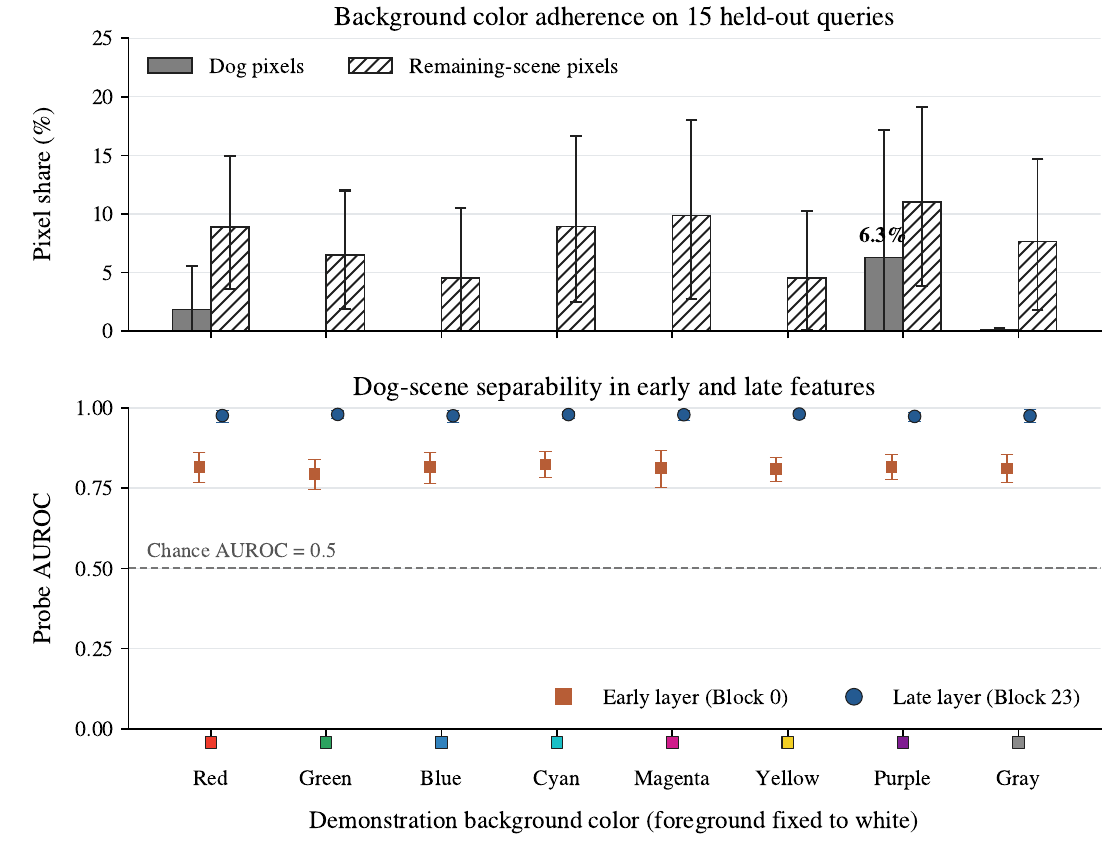}
    \caption{\textbf{Output colors and feature probes on the same 15 held-out queries.} Top: mean per-query fractions of dog and remaining-scene pixels receiving the demonstrated background color, with the foreground fixed to white. Bottom: dog-versus-scene probe AUROC for early layer and late layer features. Swatches show demonstration background colors; the dashed line marks chance AUROC (0.5). Error bars indicate 95\% bootstrap confidence intervals over queries.}    
    \label{fig:background_features}
\end{figure}

We next ask whether a region that remains black also becomes indistinguishable from the scene in the model's features. For each background color, we train separate linear probes on early features from Block 0 and late features from Block 23, using the same 15 training queries and 15 held-out queries. Figure~\ref{fig:background_features} shows that early features already distinguish the dog from the scene, while late features yield higher AUROC across all eight colors, with mean values ranging from 0.974 to 0.981. Together, the two panels show that strong linear separability in late features persists despite variation in adherence to the demonstrated background color.

The fixed person and dog example provides complementary evidence. When the non-target dog receives no magenta pixels, a probe trained on other queries still distinguishes it from the scene with AUROC $0.997$. Additional experiments examine how demonstration content affects background coloring and whether features distinguish target regions from non-target objects and the scene. These results are provided in Appendix~\ref{app:background-coloring}.

These results suggest that background coloring operates selectively over familiar semantic regions rather than uniformly over all non-target pixels. Together with the changes observed when different objects are requested, they support an account involving three learned behaviors: recognizing familiar semantic regions, interpreting demonstrations as cues for assigning target or non-target roles and their corresponding colors, and modulating these assignments through learned semantic and color preferences.

\subsection{Part-to-Parent Expansion}
\label{sec:part_expansion}

Our findings on selective background coloring also help explain part-to-parent expansion. We first examine whether the requested part remains distinguishable from the parent-only region when both receive foreground predictions. Across three part segmentation benchmarks~\citep{Gong_2017_CVPR,he2022partimagenet,wei2023ovparts}, we analyze 1,085 expansion cases from 38 tasks using binary demonstrations selected by a fixed retrieval procedure. We identify these cases by target recall $\geq 0.5$ and PL $\geq 0.25$, where recall measures the fraction of target pixels predicted as foreground and leakage measures the corresponding fraction of parent-only pixels.
Linear probes trained separately for each task on other queries distinguish the requested part from the parent-only region with AUROC averaging $0.926$ across expansion cases, and $94.6\%$ of cases reach AUROC $\geq 0.75$. 

We further examine whether late features encode which part the demonstration requests. Holding the query and demonstration images fixed, we switch the requested part and train a shared linear probe across part identities and request conditions, with cross-fitting grouped by source query image. Across 13 request pairs and 200 source query images, a region’s probe-predicted target probability increases by 0.184 when requested rather than unrequested (95\% confidence interval: [0.168, 0.201]). Among expansion requests spanning 73 unique source query images, the mean increase is 0.302 (95\% CI: $[0.271,0.332]$). Thus, information about the demonstrated target role remains readable in late features even when the final prediction expands to the parent.
Diagnostic interventions that use ground-truth masks to suppress features in the parent-only region reduce expansion and improve target IoU, whereas matched random controls yield no comparable improvement.

To investigate how background coloring affects expansion, we use 100 COCO query images with face regions defined from facial keypoints. Here, the body-only region is the person excluding the face. We keep the demonstrated face white and change every demonstration pixel outside the face from black to purple, assigning the same background color to the body-only region and the remaining scene. We group queries using only their original binary predictions. There are 56 expansion cases satisfying the criteria above and nine cases without expansion, defined by target recall $\geq 0.5$ and PL $\leq 0.1$. The remaining 35 have intermediate leakage or low recall.

\begin{figure}[t]
    \centering
    \includegraphics[width=\linewidth]{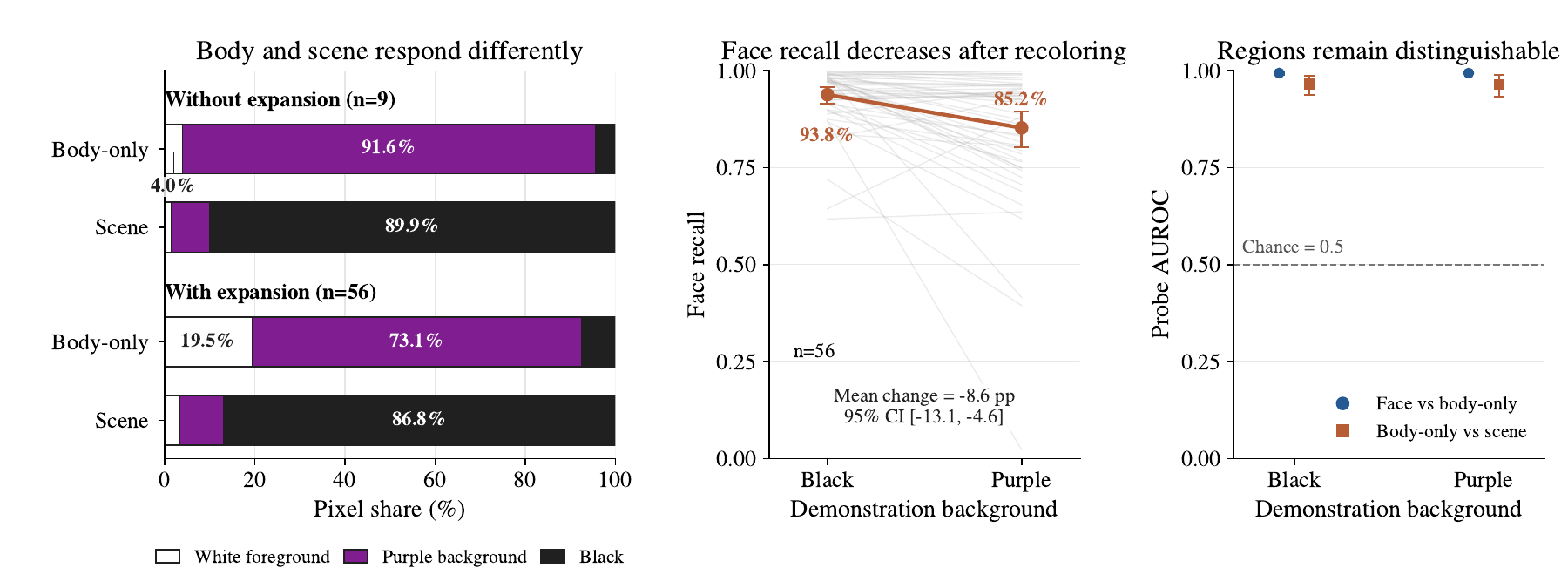}
    \caption{\textbf{Effects of changing all non-face demonstration pixels from black to purple.} Left: mean per-query output-color shares, grouped by the original binary predictions. Middle: paired face recall for the 56 expansion cases. Gray lines represent individual queries, and the colored line connects means. Right: face--body-only and body-only--scene probe AUROC under both backgrounds, with chance marked at 0.5. Error bars show 95\% bootstrap confidence intervals over queries.}
    \label{fig:face_features}
\end{figure}

Figure~\ref{fig:face_features} compares output colors, face recall, and internal distinctions under black and purple demonstration backgrounds. After changing the background to purple, body-only pixels follow the demonstrated background color more strongly in cases without expansion than in expansion cases ($91.6\%$ versus $73.1\%$). Expansion cases retain more white body pixels ($19.5\%$ versus $4.0\%$). The scene remains predominantly black, with black pixel shares of $89.9\%$ in cases without expansion and $86.8\%$ in expansion cases. The demonstrated background color therefore appears selectively on the body rather than uniformly across all regions assigned that color.

This change also affects the requested target. In the expansion group, face recall decreases from $0.938$ to $0.852$, a reduction of $8.6$ percentage points. Nevertheless, the regions remain distinguishable internally. Mean AUROC for distinguishing the face from the body-only region remains $0.993$--$0.994$ under black and purple demonstration backgrounds, while that for distinguishing the body-only region from the scene remains $0.964$--$0.966$.

The part segmentation results provide further support for this account. The requested part can remain internally distinguishable from the parent-only region even when both are predicted as foreground, and changing the demonstration background color can alter their output colors. The selective and incomplete recoloring of the body is consistent with learned semantic and color preferences influencing whether those assignments are expressed. These findings motivate explicitly marking the parent-only region in the demonstration while accounting for encoding effects on target segmentation. Appendix~\ref{app:part-expansion} provides the detailed protocols, intervention controls, and additional results.

\section{Semantic Contrast Prompt Selection (SCPS)}
Our experiments show that requested targets generally follow foreground colors once predicted as foreground, with additional palette comparisons provided in Appendix~\ref{app:foreground-colors}. Parent-only regions respond selectively to background colors, but changing the encoding can also reduce target recall. SCPS therefore combines explicit parent-only coloring with target quality calibration to select demonstrations and their encodings.

\subsection{Encoding the Target, Parent-Only Region, and Scene}
\label{sec:semantic_contrast_encoding}

Binary demonstrations assign the same background color to the parent-only region and the remaining scene. For a demonstrated target $T_d$ within a parent object $P_d$, semantic contrast assigns distinct colors to $T_d$, $P_d \setminus T_d$, and the remaining scene, explicitly marking the parent-only region for exclusion from the target. Assigning a non-black background color to all non-target pixels may also cue the coloring of other semantic regions. We therefore color only the parent-only region and keep the remaining scene black. Constructing this encoding requires target and parent annotations for the demonstration.

\begin{figure}[t]
    \centering
    \includegraphics[width=\linewidth]{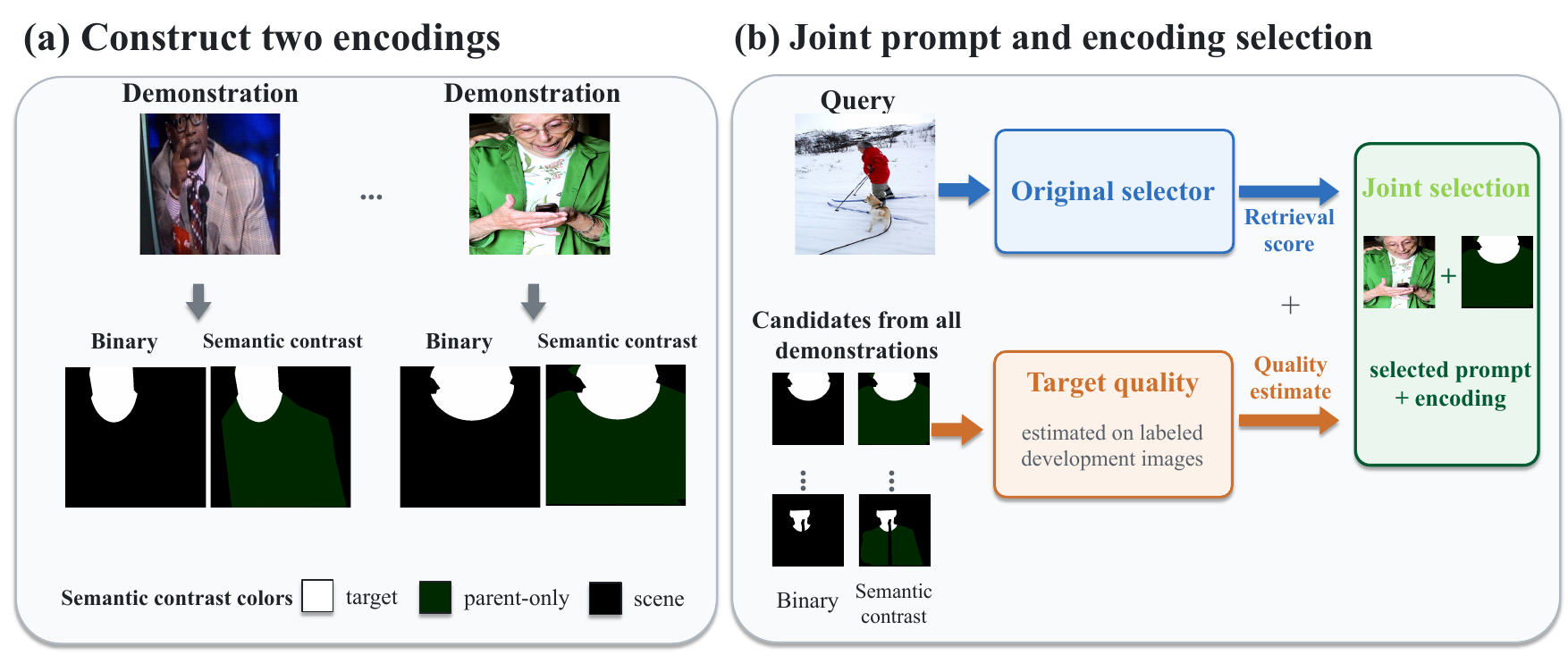}
    \caption{\textbf{Overview of Semantic Contrast Prompt Selection (SCPS).} (a) Each demonstration has binary and semantic contrast encodings. Semantic contrast assigns white to the target, dark green to the parent-only region, and black to the remaining scene. (b) SCPS combines normalized retrieval scores with weighted target quality estimates from labeled development images to jointly select a demonstration and its encoding.}
    \label{fig:scps_encoding}
\end{figure}

Figure~\ref{fig:scps_encoding}(a) illustrates the two encodings for each candidate demonstration. Both preserve the demonstration image and requested target extent. We use white for the target, dark green with normalized RGB value $(0, 0.16, 0)$ for the parent-only region, and black for the remaining scene. We select the parent-only color on the development data based on segmentation performance and fix it throughout evaluation. Only pixels assigned to white form the target mask.

\subsection{Selecting Demonstrations and Encodings}
\label{sec:scps_selection}

As illustrated in Figure~\ref{fig:scps_encoding}(b), SCPS combines retrieval scores with development estimates of target quality to select a demonstration and its encoding. We retain both binary and semantic contrast candidates. Each demonstration's original retrieval score is shared between its two encodings, while target quality is estimated separately for each version.

For each task $t$, we evaluate every candidate demonstration $p$ under both encodings on the same labeled development images. We store the mean target IoU as $\widehat{Q}_{\mathrm{target}}(t,p,m)$, where $m$ denotes the encoding.

Given a query $x_q$, let $s_{\mathrm{base}}(x_q,p)$ be the original selector's score for demonstration $p$, with larger values indicating stronger preference. We apply min--max normalization across the candidate demonstrations for each query to obtain scores in $[0,1]$, assigning zero when all scores are equal. SCPS scores each demonstration and encoding by
\begin{equation}
s_{\mathrm{SCPS}}(x_q,p,m)
=
\widetilde{s}_{\mathrm{base}}(x_q,p)
+
\alpha\,\widehat{Q}_{\mathrm{target}}(t,p,m),
\label{eq:scps_score}
\end{equation}
where $\widetilde{s}_{\mathrm{base}}$ denotes the normalized retrieval score and $\alpha$ controls the contribution of estimated target quality. We select $\alpha$ on development data separately for each retrieval method. We refer to this adjustment of the retrieval score as target quality calibration.

SCPS selects the pair $(p^*, m^*)$ with the highest score. For a fixed task and demonstration, the quality estimates determine the preferred encoding, while retrieval and quality jointly determine which demonstration is selected for the query. Using the stored estimates, SCPS requires one segmentation forward pass per new query and no test-query annotations.

\section{Evaluating SCPS}

\subsection{Experimental Setup}
\label{sec:scps_experimental_setup}

\begin{table}[t]
    \centering
    \small
    \setlength{\tabcolsep}{3pt}
    \caption{Original binary selectors and SCPS. Top: averages over 47 equally weighted tasks, with a mean across methods. Bottom: averages over tasks within each benchmark and all methods. IoU denotes target IoU, and all metrics are fractions.}
    \label{tab:main_scps}
    \label{tab:scps_datasets}
    \begin{tabular*}{\linewidth}{@{\extracolsep{\fill}}lcccccc@{}}
        \toprule
        Method / benchmark
        & \multicolumn{3}{c}{Original}
        & \multicolumn{3}{c}{SCPS} \\
        \cmidrule(lr){2-4}\cmidrule(lr){5-7}
        & IoU $\uparrow$ & PL $\downarrow$ & STS $\uparrow$
        & IoU $\uparrow$ & PL $\downarrow$ & STS $\uparrow$ \\
        \midrule
        Appearance & .2133 & .1522 & .1451 & .3037 & .1153 & .2466 \\
        CLIP~\citep{radford2021clip} & .2331 & .1596 & .1663 & .3059 & .1110 & .2514 \\
        DINOv2~\citep{oquab2023dinov2} & .2540 & .1512 & .1868 & .3140 & .1112 & .2579 \\
        UnsupPR~\citep{zhang2023goodexamples} & .2457 & .1486 & .1819 & .3104 & .1093 & .2559 \\
        SupPR~\citep{zhang2023goodexamples} & .2309 & .1512 & .1660 & .2590 & .1271 & .2055 \\
        LaPR~\citep{luo2026lapr} & .2295 & .1458 & .1702 & .2745 & .1250 & .2203 \\
        RH-Partial2Global$^{\dagger}$~\citep{wu2025rhpartial2global} & .2821 & .1495 & .2174 & .2993 & .1185 & .2419 \\
        \midrule
        Mean & .2412 & .1512 & .1762 & \textbf{.2952} & \textbf{.1168} & \textbf{.2399} \\
        \midrule[\heavyrulewidth]
        \textit{LIP} & .4026 & .0113 & .4418 & \textbf{.5199} & \textbf{.0095} & \textbf{.5880} \\
        \textit{PartImageNet-OOD} & .2229 & .2500 & .1312 & \textbf{.2471} & \textbf{.1826} & \textbf{.1673} \\
        \textit{Pascal-Part-116} & .2294 & .0823 & .1709 & \textbf{.3004} & \textbf{.0735} & \textbf{.2459} \\
        \bottomrule
    \end{tabular*}
    \par\smallskip
    \begin{minipage}{\linewidth}
        \footnotesize
        $^{\dagger}$This adapted baseline applies the reliability filter of RH-Partial2Global and then ranks demonstrations by query similarity.
    \end{minipage}
\end{table}

We evaluate SegGPT at $448 \times 448$ resolution on LIP~\citep{Gong_2017_CVPR}, PartImageNet-OOD~\citep{he2022partimagenet}, and Pascal-Part-116~\citep{wei2023ovparts}: 47 tasks and 3,313 task--query pairs from 2,336 source images. Each task provides eight demonstrations with target and parent annotations. For the seven retrieval methods in Table~\ref{tab:main_scps}, we keep each baseline's retrieval scores fixed and compare binary selection with SCPS using both encodings of the same demonstrations. Appearance ranks demonstrations by cosine similarity between globally averaged, layer-normalized SegGPT patch embeddings, without training an additional retrieval model.

On PACO-LVIS development data~\citep{Ramanathan_2023_CVPR}, we select the parent-only color by mean STS, breaking ties by higher target IoU and then lower PL, and select $\alpha$ for each retrieval method by target IoU. Both remain fixed at test time. For each task, we estimate target quality for every demonstration and encoding using three labeled development splits from its benchmark, each containing ten images, except Bicycle Seat, which has seven. We evaluate the same test queries under each split. Test source images are disjoint from development images and demonstrations.

RGB outputs are clipped to $[0,1]$ and assigned to the nearest color center by squared Euclidean distance, using white and black for binary encoding and white, dark green $(0,0.16,0)$, and black for semantic contrast. White pixels form the target mask, without spatial post-processing. We report target IoU, Parent Leakage (PL), and Strict Target Success (STS). PL is the fraction of parent-only pixels predicted as target. STS requires target IoU $\geq 0.5$, target precision $\geq 0.6$, and PL $\leq 0.1$, where precision measures the fraction of predicted target pixels belonging to the target. Benchmark results average development splits and queries within each task, then weight tasks equally.

\subsection{Main Results}
\label{sec:scps_main_results}

Table~\ref{tab:main_scps} shows gains in all three metrics for every retrieval method after aggregating benchmarks, and for every benchmark after averaging methods. Mean target IoU increases by $0.054$, PL decreases by $22.7\%$ relative to the baselines, and STS increases by $6.37$ percentage points. On LIP, baseline leakage is low, and gains mainly concern target IoU and STS.

\subsection{Ablation Studies}
\label{sec:scps_ablations}

Table~\ref{tab:scps_ablation} uses matched demonstration pools and evaluation queries. Without calibration, semantic contrast reduces PL from 0.1512 to 0.0716 but lowers target IoU, motivating quality-guided selection. Averaged across the seven retrieval methods, calibration improves target IoU and STS under both encodings, and retaining both improves these metrics over either calibrated encoding alone. Compared with calibrated binary selection, SCPS reduces PL from 0.1451 to 0.1168 and raises STS from 0.2243 to 0.2399, with higher target IoU. Paired task-bootstrap intervals support all three average gains, showing the benefits of combining semantic contrast with quality-guided selection.

SCPS retains its gains under a shared output decoder. When both candidate sets retain binary encodings and use matched calibration, semantic contrast yields lower mean PL and higher mean STS than uniform background recoloring. Appendix~\ref{app:scps-evaluation} provides the settings, detailed results, and controls.

\begin{table}[t]
    \centering
    \small
    \setlength{\tabcolsep}{6pt}
    \caption{Encoding and calibration ablations, averaged across seven retrieval methods. Calibration adds estimated target quality to retrieval scores. Bold marks the best result per metric.}
    \label{tab:scps_ablation}
    \begin{tabular}{llccc}
        \toprule
        Candidate encodings & Calibration & Target IoU $\uparrow$ & PL $\downarrow$ & STS $\uparrow$ \\
        \midrule
        Binary & No & .2412 & .1512 & .1762 \\
        Semantic contrast & No & .2214 & \textbf{.0716} & .1754 \\
        Binary & Yes & .2875 & .1451 & .2243 \\
        Semantic contrast & Yes & .2837 & .0815 & .2350 \\
        Both (SCPS) & Yes & \textbf{.2952} & .1168 & \textbf{.2399} \\
        \bottomrule
    \end{tabular}
\end{table}

\section{Discussion and Limitations}
Our findings motivate designing visual in-context segmentation models around the learned behaviors that shape demonstration following. Beyond varying output colors, training could vary whether the same semantic region is requested or excluded, and whether the target is a part or its parent. Studying these changes together may help identify and reduce dependencies among semantic identity, region role, and color that interfere with the demonstrated mapping. Evaluating internal representations alongside final predictions can reveal dependencies that output accuracy alone may obscure.

Our analysis primarily concerns SegGPT, whose RGB outputs expose both region selection and color mapping. Appendix~\ref{app:sine} provides complementary evidence from SINE's features and mask predictions, although SINE's restricted output encodings preclude the same color comparisons. Evaluating training strategies informed by these findings remains future work.

\section{Conclusion}
We identify background color mapping deviation and part-to-parent expansion in visual in-context segmentation. Across controlled changes to demonstrations, semantic regions can remain internally distinguishable while receiving colors that depart from the demonstrated mapping. These findings support an account in which demonstrations guide role and color assignments over familiar semantic regions, influenced by learned semantic and color preferences. Building on this account, SCPS combines semantic contrast encoding with target quality calibration to select a demonstration and its binary or semantic contrast encoding. Aggregated across three benchmarks, SCPS improves target IoU and STS while reducing PL for all seven evaluated retrieval methods. Our results motivate studying internal representations together with predictions when designing models that follow visual demonstrations more reliably.

\subsection*{AI use statement}
The authors conducted the initial experiments and observations, iteratively formulated, tested, and refined the hypotheses, and designed and performed the main experiments supporting the proposed mechanistic account. Generative AI tools helped suggest supplementary validation experiments, write experimental and analysis scripts, and discuss interpretations of the results. We also used these tools to polish the writing and identify relevant literature, including recent related work we had missed. The authors reviewed the AI-assisted contributions and take responsibility for the final content, results, and claims.

\bibliography{references}

@inproceedings{min2022rethinking,
    title = "Rethinking the Role of Demonstrations: What Makes In-Context Learning Work?",
    author = "Min, Sewon  and
      Lyu, Xinxi  and
      Holtzman, Ari  and
      Artetxe, Mikel  and
      Lewis, Mike  and
      Hajishirzi, Hannaneh  and
      Zettlemoyer, Luke",
    editor = "Goldberg, Yoav  and
      Kozareva, Zornitsa  and
      Zhang, Yue",
    booktitle = "Proceedings of the 2022 Conference on Empirical Methods in Natural Language Processing",
    month = dec,
    year = "2022",
    address = "Abu Dhabi, United Arab Emirates",
    publisher = "Association for Computational Linguistics",
    url = "https://aclanthology.org/2022.emnlp-main.759/",
    doi = "10.18653/v1/2022.emnlp-main.759",
    pages = "11048--11064"
}

@inproceedings{pan2023context,
    title = "What In-Context Learning ``Learns'' In-Context: Disentangling Task Recognition and Task Learning",
    author = "Pan, Jane  and
      Gao, Tianyu  and
      Chen, Howard  and
      Chen, Danqi",
    editor = "Rogers, Anna  and
      Boyd-Graber, Jordan  and
      Okazaki, Naoaki",
    booktitle = "Findings of the Association for Computational Linguistics: ACL 2023",
    month = jul,
    year = "2023",
    address = "Toronto, Canada",
    publisher = "Association for Computational Linguistics",
    url = "https://aclanthology.org/2023.findings-acl.527/",
    doi = "10.18653/v1/2023.findings-acl.527",
    pages = "8298--8319"
}

@inproceedings{NEURIPS2022_9f09f316,
 author = {Bar, Amir and Gandelsman, Yossi and Darrell, Trevor and Globerson, Amir and Efros, Alexei},
 booktitle = {Advances in Neural Information Processing Systems},
 doi = {10.52202/068431-1813},
 editor = {S. Koyejo and S. Mohamed and A. Agarwal and D. Belgrave and K. Cho and A. Oh},
 pages = {25005--25017},
 publisher = {Curran Associates, Inc.},
 title = {Visual Prompting via Image Inpainting},
 url = {https://proceedings.neurips.cc/paper_files/paper/2022/file/9f09f316a3eaf59d9ced5ffaefe97e0f-Paper-Conference.pdf},
 volume = {35},
 year = {2022}
}

@misc{liao2026enhancingvisualincontextlearning,
      title={Enhancing Visual In-Context Learning by Multi-Faceted Fusion}, 
      author={Wenwen Liao and Jianbo Yu and Yuansong Wang and Qingchao Jiang and Xiaofeng Yang},
      year={2026},
      eprint={2601.10107},
      archivePrefix={arXiv},
      primaryClass={cs.CV},
      howpublished={arXiv:2601.10107},
      url={https://arxiv.org/abs/2601.10107}, 
}

@InProceedings{Wang_2023_CVPR,
    author    = {Wang, Xinlong and Wang, Wen and Cao, Yue and Shen, Chunhua and Huang, Tiejun},
    title     = {Images Speak in Images: A Generalist Painter for In-Context Visual Learning},
    booktitle = {Proceedings of the IEEE/CVF Conference on Computer Vision and Pattern Recognition (CVPR)},
    month     = {June},
    year      = {2023},
    pages     = {6830-6839}
}

@InProceedings{Bai_2024_CVPR,
    author    = {Bai, Yutong and Geng, Xinyang and Mangalam, Karttikeya and Bar, Amir and Yuille, Alan L. and Darrell, Trevor and Malik, Jitendra and Efros, Alexei A.},
    title     = {Sequential Modeling Enables Scalable Learning for Large Vision Models},
    booktitle = {Proceedings of the IEEE/CVF Conference on Computer Vision and Pattern Recognition (CVPR)},
    month     = {June},
    year      = {2024},
    pages     = {22861-22872}
}

@InProceedings{pmlr-v235-hao24b,
  title = 	 {Data-efficient Large Vision Models through Sequential Autoregression},
  author =       {Hao, Zhiwei and Guo, Jianyuan and Wang, Chengcheng and Tang, Yehui and Wu, Han and Hu, Han and Han, Kai and Xu, Chang},
  booktitle = 	 {Proceedings of the 41st International Conference on Machine Learning},
  pages = 	 {17572--17596},
  year = 	 {2024},
  editor = 	 {Salakhutdinov, Ruslan and Kolter, Zico and Heller, Katherine and Weller, Adrian and Oliver, Nuria and Scarlett, Jonathan and Berkenkamp, Felix},
  volume = 	 {235},
  series = 	 {Proceedings of Machine Learning Research},
  month = 	 {21--27 Jul},
  publisher =    {PMLR},
  url = 	 {https://proceedings.mlr.press/v235/hao24b.html}
}

@InProceedings{VTC,
author="Hojel, Alberto
and Bai, Yutong
and Darrell, Trevor
and Globerson, Amir
and Bar, Amir",
editor="Leonardis, Ale{\v{s}}
and Ricci, Elisa
and Roth, Stefan
and Russakovsky, Olga
and Sattler, Torsten
and Varol, G{\"u}l",
title="Finding Visual Task Vectors",
booktitle="Computer Vision -- ECCV 2024",
year="2025",
publisher="Springer Nature Switzerland",
address="Cham",
pages="257--273",
doi="10.1007/978-3-031-72775-7_15",
isbn="978-3-031-72775-7"
}

@misc{khatri2026modelsizeprobinggaps,
      title={Beyond Model Size: Probing the Gaps in Visual in-Context Learning by Training a Tiny Model}, 
      author={Sunil Khatri and Steven Landgraf and Markus Ulrich and Simon Rei{\ss}},
      year={2026},
      eprint={2606.10905},
      archivePrefix={arXiv},
      primaryClass={cs.CV},
      howpublished={arXiv:2606.10905},
      url={https://arxiv.org/abs/2606.10905}, 
}

@inproceedings{NEURIPS2024_2cc0b084,
 author = {Liu, Yang and Jing, Chenchen and Li, Hengtao and Zhu, Muzhi and Chen, Hao and Wang, Xinlong and Shen, Chunhua},
 booktitle = {Advances in Neural Information Processing Systems},
 doi = {10.52202/079017-0791},
 editor = {A. Globerson and L. Mackey and D. Belgrave and A. Fan and U. Paquet and J. Tomczak and C. Zhang},
 pages = {25095--25119},
 publisher = {Curran Associates, Inc.},
 title = {A Simple Image Segmentation Framework via In-Context Examples},
 url = {https://proceedings.neurips.cc/paper_files/paper/2024/file/2cc0b08447bf9668db268e6c86364a6e-Paper-Conference.pdf},
 volume = {37},
 year = {2024}
}

@InProceedings{Wang_2023_ICCV,
    author    = {Wang, Xinlong and Zhang, Xiaosong and Cao, Yue and Wang, Wen and Shen, Chunhua and Huang, Tiejun},
    title     = {{SegGPT}: Towards Segmenting Everything in Context},
    booktitle = {Proceedings of the IEEE/CVF International Conference on Computer Vision (ICCV)},
    month     = {October},
    year      = {2023},
    pages     = {1130-1140}
}

@InProceedings{pmlr-v235-liu24k,
  title = 	 {Unifying Image Processing as Visual Prompting Question Answering},
  author =       {Liu, Yihao and Chen, Xiangyu and Ma, Xianzheng and Wang, Xintao and Zhou, Jiantao and Qiao, Yu and Dong, Chao},
  booktitle = 	 {Proceedings of the 41st International Conference on Machine Learning},
  pages = 	 {30873--30891},
  year = 	 {2024},
  editor = 	 {Salakhutdinov, Ruslan and Kolter, Zico and Heller, Katherine and Weller, Adrian and Oliver, Nuria and Scarlett, Jonathan and Berkenkamp, Felix},
  volume = 	 {235},
  series = 	 {Proceedings of Machine Learning Research},
  month = 	 {21--27 Jul},
  publisher =    {PMLR},
  url = 	 {https://proceedings.mlr.press/v235/liu24k.html}
}

@inproceedings{chen2024learning,
  title={Learning A Low-Level Vision Generalist via Visual Task Prompt},
  author={Xiangyu Chen and Yihao Liu and Yuandong Pu and Wenlong Zhang and Jiantao Zhou and Yu Qiao and Chao Dong},
  booktitle={Proceedings of the 32nd ACM International Conference on Multimedia},
  year={2024},
  pages={2671--2680},
  doi={10.1145/3664647.3681621}
}

@inproceedings{Rajagopalan2024AWRaCLeAI,
  title={{AWRaCLe}: All-Weather Image Restoration Using Visual In-Context Learning},
  author={Rajagopalan, Sudarshan and Patel, Vishal M.},
  booktitle={Proceedings of the AAAI Conference on Artificial Intelligence},
  volume={39},
  pages={6675--6683},
  year={2025},
  doi={10.1609/aaai.v39i6.32716},
  url={https://ojs.aaai.org/index.php/AAAI/article/view/32716}
}

@InProceedings{radford2021clip,
  title = 	 {Learning Transferable Visual Models From Natural Language Supervision},
  author =       {Radford, Alec and Kim, Jong Wook and Hallacy, Chris and Ramesh, Aditya and Goh, Gabriel and Agarwal, Sandhini and Sastry, Girish and Askell, Amanda and Mishkin, Pamela and Clark, Jack and Krueger, Gretchen and Sutskever, Ilya},
  booktitle = 	 {Proceedings of the 38th International Conference on Machine Learning},
  pages = 	 {8748--8763},
  year = 	 {2021},
  editor = 	 {Meila, Marina and Zhang, Tong},
  volume = 	 {139},
  series = 	 {Proceedings of Machine Learning Research},
  month = 	 {18--24 Jul},
  publisher =    {PMLR},
  url = 	 {https://proceedings.mlr.press/v139/radford21a.html}
}

@misc{oquab2023dinov2,
  title={{DINOv2}: Learning Robust Visual Features without Supervision},
  author={Oquab, Maxime and Darcet, Timoth\'ee and Moutakanni, Th\'eo and
          Vo, Huy and Szafraniec, Marc and Khalidov, Vasil and Fernandez, Pierre and
          Haziza, Daniel and Massa, Francisco and El-Nouby, Alaaeldin and
          Assran, Mahmoud and Ballas, Nicolas and Galuba, Wojciech and
          Howes, Russell and Huang, Po-Yao and Li, Shang-Wen and Misra, Ishan and
          Rabbat, Michael and Sharma, Vasu and Synnaeve, Gabriel and Xu, Hu and
          Jegou, Herv\'e and Mairal, Julien and Labatut, Patrick and Joulin, Armand and
          Bojanowski, Piotr},
  howpublished={arXiv:2304.07193},
  url={https://arxiv.org/abs/2304.07193v1},
  year={2023}
}

@inproceedings{zhang2023goodexamples,
 author = {Zhang, Yuanhan and Zhou, Kaiyang and Liu, Ziwei},
 booktitle = {Advances in Neural Information Processing Systems},
 doi = {10.52202/075280-0780},
 editor = {A. Oh and T. Naumann and A. Globerson and K. Saenko and M. Hardt and S. Levine},
 pages = {17773--17794},
 publisher = {Curran Associates, Inc.},
 title = {What Makes Good Examples for Visual In-Context Learning?},
 url = {https://proceedings.neurips.cc/paper_files/paper/2023/file/398ae57ed4fda79d0781c65c926d667b-Paper-Conference.pdf},
 volume = {36},
 year = {2023}
}

@InProceedings{luo2026lapr,
    author    = {Luo, Tianci and Pan, Haohao and Wang, Jinpeng and Lian, Niu and Chen, Xinrui and Chen, Bin and Xia, Shu-Tao and Yuan, Chun},
    title     = {Love Me, Love My Label: Rethinking the Role of Labels in Prompt Retrieval for Visual In-Context Learning},
    booktitle = {Proceedings of the IEEE/CVF Conference on Computer Vision and Pattern Recognition (CVPR)},
    month     = {June},
    year      = {2026},
    pages     = {38774-38783}
}

@inproceedings{xu2024partial2global,
 author = {Xu, Chengming and Liu, Chen and Wang, Yikai and Yao, Yuan and Fu, Yanwei},
 booktitle = {Advances in Neural Information Processing Systems},
 doi = {10.52202/079017-2384},
 editor = {A. Globerson and L. Mackey and D. Belgrave and A. Fan and U. Paquet and J. Tomczak and C. Zhang},
 pages = {74945--74965},
 publisher = {Curran Associates, Inc.},
 title = {Towards Global Optimal Visual In-Context Learning Prompt Selection},
 url = {https://proceedings.neurips.cc/paper_files/paper/2024/file/8900e600e4f55be0a10cee4a14d7c3de-Paper-Conference.pdf},
 volume = {37},
 year = {2024}
}

@inproceedings{wu2025rhpartial2global,
  title={Towards Reliable and Holistic Visual In-Context Learning Prompt Selection},
  author={Wenxiao Wu and Jing-Hao Xue and Chengming Xu and Chen Liu and Xinwei Sun and Changxin Gao and Nong Sang and Yanwei Fu},
  booktitle={Advances in Neural Information Processing Systems},
  volume={38},
  pages={123336--123356},
  year={2025},
  doi={10.52202/085713-3715},
  url={https://proceedings.neurips.cc/paper_files/paper/2025/hash/a17251f8d595179eef5e466b1f5f7a85-Abstract-Conference.html}
}

@InProceedings{Ramanathan_2023_CVPR,
    author    = {Ramanathan, Vignesh and Kalia, Anmol and Petrovic, Vladan and Wen, Yi and Zheng, Baixue and Guo, Baishan and Wang, Rui and Marquez, Aaron and Kovvuri, Rama and Kadian, Abhishek and Mousavi, Amir and Song, Yiwen and Dubey, Abhimanyu and Mahajan, Dhruv},
    title     = {{PACO}: Parts and Attributes of Common Objects},
    booktitle = {Proceedings of the IEEE/CVF Conference on Computer Vision and Pattern Recognition (CVPR)},
    month     = {June},
    year      = {2023},
    pages     = {7141-7151}
}

@InProceedings{Gong_2017_CVPR,
author = {Gong, Ke and Liang, Xiaodan and Zhang, Dongyu and Shen, Xiaohui and Lin, Liang},
title = {Look Into Person: Self-Supervised Structure-Sensitive Learning and a New Benchmark for Human Parsing},
booktitle = {Proceedings of the IEEE Conference on Computer Vision and Pattern Recognition (CVPR)},
month = {July},
year = {2017},
pages = {6757--6765},
doi = {10.1109/CVPR.2017.715}
}

@InProceedings{he2022partimagenet,
author="He, Ju
and Yang, Shuo
and Yang, Shaokang
and Kortylewski, Adam
and Yuan, Xiaoding
and Chen, Jie-Neng
and Liu, Shuai
and Yang, Cheng
and Yu, Qihang
and Yuille, Alan",
editor="Avidan, Shai
and Brostow, Gabriel
and Ciss{\'e}, Moustapha
and Farinella, Giovanni Maria
and Hassner, Tal",
title="{PartImageNet}: A Large, High-Quality Dataset of Parts",
booktitle="Computer Vision -- ECCV 2022",
year="2022",
publisher="Springer Nature Switzerland",
address="Cham",
pages="128--145",
doi="10.1007/978-3-031-20074-8_8",
isbn="978-3-031-20074-8"
}

@inproceedings{wei2023ovparts,
 author = {Wei, Meng and Yue, Xiaoyu and Zhang, Wenwei and Kong, Shu and Liu, Xihui and Pang, Jiangmiao},
 booktitle = {Advances in Neural Information Processing Systems},
 doi = {10.52202/075280-3071},
 editor = {A. Oh and T. Naumann and A. Globerson and K. Saenko and M. Hardt and S. Levine},
 pages = {70094--70114},
 publisher = {Curran Associates, Inc.},
 title = {{OV-PARTS}: Towards Open-Vocabulary Part Segmentation},
 url = {https://proceedings.neurips.cc/paper_files/paper/2023/file/dde53059fdb0f45e1e9ad9c66997d662-Paper-Datasets_and_Benchmarks.pdf},
 volume = {36},
 year = {2023}
}

@InProceedings{lin2014microsoft,
author="Lin, Tsung-Yi
and Maire, Michael
and Belongie, Serge
and Hays, James
and Perona, Pietro
and Ramanan, Deva
and Doll{\'a}r, Piotr
and Zitnick, C. Lawrence",
editor="Fleet, David
and Pajdla, Tomas
and Schiele, Bernt
and Tuytelaars, Tinne",
title="Microsoft {COCO}: Common Objects in Context",
booktitle="Computer Vision -- ECCV 2014",
year="2014",
publisher="Springer International Publishing",
address="Cham",
pages="740--755",
doi="10.1007/978-3-319-10602-1_48",
isbn="978-3-319-10602-1"
}

@misc{autodistill2023seggpt,
  title     = {{SegGPT} Base Models Documentation},
  author    = {{Autodistill Team}},
  year      = {2023},
  url       = {https://docs.autodistill.com/base_models/seggpt/},
  note      = {Accessed: 2025-08-18}
}
\bibliographystyle{preprint_refs}

\appendix
\section{Appendix}
\subsection{Additional analysis details}
\label{app:analysis-details}

We provide analysis protocols, additional evidence on background coloring and
part-to-parent expansion, SCPS settings and controls, and a complementary
analysis of SINE.

\subsubsection{Color comparisons and region definitions}
\label{app:color-definitions}

The background experiment uses one demonstration requesting the person and
30 COCO queries. The foreground stays white while the background takes one
of eight colors: red, green, blue, cyan, magenta, yellow, purple, or gray.
Their normalized RGB values are $(1,0,0)$, $(0,1,0)$, $(0,0,1)$,
$(0,1,1)$, $(1,0,1)$, $(1,1,0)$, $(0.5,0,0.5)$, and
$(0.5,0.5,0.5)$, respectively. The demonstration image, target mask,
and query remain fixed across colors.

Color adherence is the fraction of pixels in an annotated query region
assigned to its demonstrated color. We assign predictions to the nearest
demonstrated foreground color, background color, or black by RGB distance.
In the person, dog, and horse analyses, the remaining scene excludes those
annotated objects but may contain other semantic regions, such as sky or
vegetation. In part segmentation, the parent-only region is the parent object
excluding the requested part, and the remaining scene lies outside the parent.

\subsubsection{Feature probes}
\label{app:feature-probes}

Unless otherwise specified, we extract SegGPT query features from the final transformer block, Block 23, after the model's final normalization, during the same forward pass as the RGB prediction. For the early-layer comparison in Figure~\ref{fig:background_features}, we additionally extract query features from the output of the first transformer block, Block 0. A token's region occupancy is the fraction of its image patch covered by that region.

For background coloring, dog and scene tokens require occupancies of at least 0.25 and 0.80, respectively. We train a separate linear probe for each layer and background color using the same 15 training queries and 15 evaluation queries. We compute AUROC separately for each evaluation query and report the mean across queries.

For part segmentation, target and parent-only tokens require occupancies of
at least 0.25 and 0.50, respectively; tokens meeting both criteria are excluded.
We normalize features to unit L2 norm and fit linear logistic probes separately
for each task. Queries are divided into three fixed folds, and each query is
evaluated with a probe trained on the other two folds. Probes use AdamW for
300 steps with learning rate 0.02 and weight decay $10^{-4}$, using at most
12,000 training tokens per class and 1,024 tokens per class in each batch.

For face segmentation, occupancy thresholds are 0.25 for the face, 0.50 for
the body-only region, and 0.80 for the scene. Tokens meeting multiple
thresholds are excluded. We use five folds grouped by source image and
evaluate each image with probes trained on other images. Separate probes are
fitted under black and purple backgrounds to test whether the annotated
distinctions remain readable under each encoding.

The shared part-role probe uses the paired-request protocol described in
Appendix~\ref{app:part-role}.

\subsubsection{Averaging and confidence intervals}
\label{app:averaging}

We compute color shares within each query before averaging across queries.
Paired bootstrap comparisons resample query identities together across
conditions; the comparison of requested objects uses 10,000 resamples.
For SCPS, we average queries and development splits within each task, then
resample task means. Feature intervention intervals resample tasks while
retaining all sampled cases within each task. For the shared part-role probe,
we average repeated part-pair comparisons within each source query image
before bootstrapping source query images. All confidence intervals are 95\%
intervals.

\subsection{Additional evidence on background coloring}
\label{app:background-coloring}

\subsubsection{Background colors and feature separability}
\label{app:background-separability}

Across all 30 queries, purple gives the highest mean dog adherence at 17.3\%.
In six queries, more than half of the dog pixels receive purple; no query
reaches this level under the other seven colors. Figure~\ref{fig:background_features} in the main paper
uses the 15 held-out queries evaluated by the probes, where purple adherence
averages 6.3\%. Table~\ref{tab:app-dog-probes} shows that the dog remains distinguishable
from the scene under every tested color.

\begin{table}[!htbp]
\centering
\caption{Dog-versus-scene probes using late features from Block 23, fitted separately for each demonstration background color. Mean AUROC and 95\% bootstrap confidence intervals are computed over the same 15 held-out queries used in Figure~\ref{fig:background_features}.}
\label{tab:app-dog-probes}
\begin{tabular}{lcc}
\hline
Background & Mean AUROC & 95\% CI \\
\hline
Red     & 0.9760 & $[0.9542,0.9930]$ \\
Green   & 0.9798 & $[0.9657,0.9917]$ \\
Blue    & 0.9755 & $[0.9541,0.9924]$ \\
Cyan    & 0.9792 & $[0.9667,0.9901]$ \\
Magenta & 0.9787 & $[0.9614,0.9929]$ \\
Yellow  & 0.9810 & $[0.9658,0.9932]$ \\
Purple  & 0.9736 & $[0.9577,0.9870]$ \\
Gray    & 0.9753 & $[0.9537,0.9939]$ \\
\hline
\end{tabular}
\end{table}

\paragraph{Unit RGB-direction projection.}
We additionally compare continuous RGB predictions under purple
$(0.5,0,0.5)$ and magenta $(1,0,1)$. For query $i$ and demonstration
background color $c$, we compute the unit RGB-direction projection
\begin{equation}
s_{ic} =
\left(\frac{1}{|D_i|}\sum_{p\in D_i}\mathbf{y}_{ic}(p)\right)^{\!\top}
\frac{\mathbf{b}_c}{\|\mathbf{b}_c\|_2},
\label{eq:app-rgb-projection}
\end{equation}
where $D_i$ is the annotated dog region, $\mathbf{y}_{ic}(p)$ is the
floating-point RGB output before saving the outputs as PNG images or applying nearest-color decoding, and $\mathbf{b}_c$ is the demonstration background color.
Purple and magenta have the same unit RGB direction,
$(1/\sqrt{2},0,1/\sqrt{2})$, while differing in magnitude.
Across the 30 queries, mean $s_{ic}$ is 0.1175 under purple and 0.0100
under magenta. The mean paired purple-minus-magenta difference is 0.1075,
with a query-bootstrap 95\% confidence interval of $[0.0385,0.1853]$.
The color dependence is therefore present in continuous predictions and
is not solely induced by nearest-color decoding.

\subsubsection{Foreground colors and target recovery}
\label{app:foreground-colors}

We vary the foreground over red, green, blue, cyan, magenta, yellow, white,
and gray under black or purple backgrounds, producing 480 outputs on the
same 30 queries. To distinguish the requested color from other colors, we
assign predictions to the nearest of the eight foreground colors, the
demonstrated background color, and black.

Table~\ref{tab:app-foreground-colors} reports the fraction of annotated
target pixels receiving the requested color and target IoU using decoding
with the demonstrated colors and black. Gray exceeds red in color share by
7.1 percentage points under black backgrounds and 14.2 points under purple
backgrounds, with paired intervals $[4.5,10.4]$ and $[10.4,18.9]$.
Among target pixels predicted as foreground, agreement with the demonstrated
foreground color ranges from 95.1\% to 100\%. Foreground predictions
therefore generally follow the specified color, while the recovered target
extent varies with the palette.

\begin{table}[!htbp]
\centering
\caption{Foreground color comparisons on 30 queries. Shares of the requested
color use the expanded palette described in the text and are percentages of
annotated target pixels. Target IoU uses decoding with the demonstrated
colors and black.}
\label{tab:app-foreground-colors}
\begin{tabular}{lrrrr}
\hline
& \multicolumn{2}{c}{Black background}
& \multicolumn{2}{c}{Purple background} \\
\cline{2-3}\cline{4-5}
Foreground & Color share (\%) & Target IoU & Color share (\%) & Target IoU \\
\hline
Red     & 73.5 & .663 & 66.4 & .615 \\
Green   & 76.6 & .681 & 74.6 & .672 \\
Blue    & 78.0 & .687 & 74.7 & .670 \\
Cyan    & 78.6 & .696 & 76.9 & .685 \\
Magenta & 79.4 & .703 & 74.2 & .665 \\
Yellow  & 78.9 & .700 & 76.4 & .687 \\
White   & 78.4 & .700 & 75.2 & .684 \\
Gray    & 80.6 & .701 & 80.6 & .699 \\
\hline
\end{tabular}
\end{table}

\subsubsection{Changing the requested object across queries}
\label{app:requested-object}

We extend the person and dog comparison to 30 queries, three demonstrations,
and four color pairs fixed before the experiment. Demonstration images are
disjoint from the queries. The person and dog are assigned, respectively,
gray and magenta, white and purple, cyan and red, or yellow and blue.

We request each object under every demonstration–query and color-pair combination, producing 720 outputs in total. The
requested object receives its assigned color and all other pixels receive
the other object's color. For example, requesting the person gives gray
person and magenta background, while requesting the dog gives magenta dog
and gray background. Each object's assigned RGB value stays fixed while
its role and the scene's background color change. Predictions are decoded
using the two demonstrated colors and black.

We average the 12 demonstration and color combinations within each query.
Table~\ref{tab:app-object-roles} reports adherence under each role and the
paired difference. Requesting the dog increases adherence to its assigned
color, whereas requesting the person produces a smaller decrease.

The original fixed demonstration and query pair gives mean increases of
76.7 percentage points for the dog and 4.2 points for the person across
42 color pairs. The broader comparison confirms a positive mean effect for
the dog while showing that the size and direction of the effect depend on
the semantic region and image conditions.

\begin{table}[!htbp]
\centering
\caption{Color adherence across 30 queries. Adherence is the percentage of
the annotated region receiving its specified color. Differences and their
query bootstrap intervals are in percentage points.}
\label{tab:app-object-roles}
\begin{tabular}{lrrc}
\hline
Region & Requested & Not requested & Difference [95\% confidence interval] \\
\hline
Person & 73.0 & 75.5 & $-2.5\;[-3.9,-1.1]$ \\
Dog    & 54.1 & 33.0 & $+21.1\;[15.2,26.6]$ \\
\hline
\end{tabular}
\end{table}

\subsubsection{Information about target and non-target roles}
\label{app:object-role-probe}

A separate linear classifier predicts whether a token belongs to the target,
a non-target object, or the scene. We train it on 80 queries with requests
for both person and dog and evaluate it on 48 disjoint queries with a
held-out demonstration. Balanced accuracy is 0.929. When the person becomes
requested, its mean predicted target probability increases by 0.878; the
corresponding increase for the dog is 0.820. Internal features thus contain
information about the demonstrated role as well as semantic region identity.

\subsubsection{Demonstration content and region coloring}
\label{app:demonstration-content}

We compare two natural demonstrations, one containing a horse and one without
a horse, on the same 30 queries containing a person, dog, and horse. Both
demonstrations request the person. The query horse receives the demonstrated
background color on 89.9\% of its pixels with the horse-containing
demonstration and 0.2\% with the other demonstration, a paired difference
of 89.7 percentage points with interval $[84.8,93.6]$.

Probes trained on 15 queries and evaluated on the other 15 distinguish the
horse from the scene with mean AUROC 0.977 and 0.974 under the respective
demonstrations. Transferring the first probe to the second condition gives
AUROC 0.918; the reverse transfer gives 0.972. Output coloring changes while
the horse remains internally distinguishable. Because the demonstrations
also differ in other image content, this comparison measures the effect of
changing the demonstration as a whole.

For the fixed person and dog example, we also separate the demonstrated dog
from the scene. With gray person and magenta background, the query person
is 94.9\% gray and the dog receives no magenta pixels. Keeping the
demonstrated person gray and dog magenta while changing the scene to black
yields 94.6\% gray person pixels, 89.2\% magenta dog pixels, and 99.2\%
black scene pixels. Explicitly distinguishing the dog from the scene changes
whether its assigned color appears in the prediction.

\subsection{Additional evidence on part-to-parent expansion}
\label{app:part-expansion}

\subsubsection{Part separability across queries}
\label{app:part-separability}

The part analysis uses one selected binary prediction for each of the 3,313 task--query pairs. A ridge regressor trained on PACO-LVIS development data predicts target IoU from one patch-embedding appearance similarity and 12 Block-23 prompt--query compatibility features. The compatibility features use demonstration target and parent annotations to construct regional prototypes, without using query annotations. For each query, we select the highest-scoring demonstration from eight candidates and analyze its precomputed binary prediction. This selector fixes demonstration choices for the part analyses and is separate from the seven retrieval methods evaluated in Table~\ref{tab:main_scps}.

Target recall $\geq 0.5$ and PL $\geq 0.25$ identify 1,085 expansion
cases from 38 tasks. Probes trained on other queries achieve mean AUROC
0.926 across these cases, with 94.6\% reaching AUROC $\geq 0.75$.
With equal task weights, mean AUROC is 0.916, and the mean fraction reaching
AUROC $\geq 0.75$ is 91.7\%.

\subsubsection{Face masks and paired recoloring}
\label{app:face-recoloring}

Face masks are ellipses derived from five visible COCO facial keypoints and
intersected with the annotated person mask. The body-only region is the
person excluding the face. The demonstration uses image ID 191577 and
person annotation ID 1723055. Its face stays white while all other pixels
change from black to purple $(0.5,0,0.5)$.

Binary predictions define the groups: 56 expansion cases, nine cases without
expansion, and 35 with intermediate leakage or low recall.
Table~\ref{tab:app-face-recoloring} reports the recoloring results.
Under purple backgrounds, body adherence is 18.6 percentage points higher
in cases without expansion than in expansion cases, with interval
$[11.6,25.8]$. Face recall in the expansion group changes by $-8.6$
percentage points, with paired interval $[-13.1,-4.6]$.

\begin{table}[!htbp]
\centering
\small
\caption{Recoloring all non-face demonstration pixels purple. Groups are
defined by binary predictions. Color shares are percentages; face recall
is a fraction.}
\label{tab:app-face-recoloring}
\begin{tabular}{lrr}
\hline
Measure & Without expansion ($n=9$) & With expansion ($n=56$) \\
\hline
Body-only purple share       & 91.6 & 73.1 \\
Body-only white share        & 4.0  & 19.5 \\
Body-only black share        & 4.4  & 7.5 \\
Scene purple share           & 8.7  & 9.9 \\
Scene black share            & 89.9 & 86.8 \\
\hline
Face recall, black background  & .785 & .938 \\
Face recall, purple background & .834 & .852 \\
\hline
\end{tabular}
\end{table}

\subsubsection{Diagnostic feature interventions}
\label{app:feature-interventions}

We sample 200 of the 1,085 expansion cases in
Appendix~\ref{app:part-separability}, stratifying by task. The sample covers
38 tasks. All conditions keep the query, selected demonstration, and binary
encoding fixed.

We modify query features entering the decoder embedding. Let $T$, $R$,
and $S$ denote token sets for the target, parent-only region, and remaining
scene, with mean features $\mu_T$, $\mu_R$, and $\mu_S$.
Target enhancement and parent-only suppression update token feature $z_i$ by
\[
\begin{array}{rcll}
z_i' & = & z_i + \lambda_T(\mu_T-\mu_R), & i\in T, \\
z_i' & = & (1-\lambda_R)z_i + \lambda_R\mu_S, & i\in R,
\end{array}
\]
where $\lambda_T=\lambda_R=0.25$. The first adds one quarter of the
difference between the target and parent-only mean features; the second
moves each parent-only token one quarter of the way toward the scene mean.
We evaluate them separately and together. Ground-truth masks define the token sets and the diagnostic updates.

Random direction controls use updates with matched norms at the original
locations. The random location control applies both components to two
disjoint sets sampled from the full query token grid, matching the target
and parent-only token counts. It preserves the target update; the parent
component follows $\mu_S-\mu_R$, scaled to the mean norm of the original
parent-only updates. Sampled locations may overlap the annotated regions
by chance.

Table~\ref{tab:app-interventions} shows that parent-only suppression
substantially reduces leakage and improves target IoU. Target enhancement
alone changes little, and random controls yield no comparable improvement.
Using 10,000 paired bootstrap resamples of the 38 tasks, the target IoU
increase from parent-only suppression is 0.4008, with interval
$[0.3710,0.4297]$. The change in PL is $-0.6556$, with interval
$[-0.7122,-0.5957]$. Target recall changes by $-0.0560$, with interval
$[-0.0654,-0.0474]$. Parent-only features contribute to the erroneous
foreground prediction, although suppressing them also removes some target
pixels.

\begin{table}[!htbp]
\centering
\caption{Feature interventions on 200 expansion cases. Values are averaged
across cases. Ground-truth masks are used only for this diagnostic, not by
SCPS at test time.}
\label{tab:app-interventions}
\begin{tabular}{lrrr}
\hline
Condition & Target IoU & Target recall & PL \\
\hline
Unmodified prediction         & .2054 & .8553 & .6916 \\
Target enhancement            & .2073 & .8569 & .6923 \\
Parent-only suppression       & .6063 & .7993 & .0360 \\
Both interventions            & .6062 & .8040 & .0372 \\
Random target directions      & .2052 & .8537 & .6916 \\
Random parent-only directions & .2058 & .8551 & .6902 \\
Both random controls          & .2059 & .8544 & .6894 \\
Random location control       & .1939 & .7166 & .5811 \\
\hline
\end{tabular}
\end{table}

\subsubsection{Separability and the response to semantic contrast}
\label{app:separability-response}

Among the 1,085 expansion cases, we compare binary and semantic contrast encodings of the same demonstration. PL decreases without reducing target IoU in 46.9\% of cases with probe AUROC $\geq 0.75$ and 21.2\% of the other cases, using equal task weights. Across the 20 tasks containing both groups, the paired difference is 25.4 percentage points with interval $[9.9,38.7]$. Greater part separability is associated with a more favorable response to semantic contrast.

\subsubsection{Demonstration-conditioned part-role information}
\label{app:part-role}

We examine whether late features encode the requested role of a part by pairing demonstrations that request different parts of the same parent. Within each pair, the query image and demonstration image remain fixed, while the demonstration target mask changes. Paired demonstrations must have parent-mask IoU of at least 0.90 and requested-part IoU of at most 0.10. The evaluation contains 13 request pairs, 305 paired query instances, and 200 unique source query images.

We fit one shared linear probe across part identities and both request
conditions. Five-fold cross-fitting groups instances by dataset and source
query image, keeping all request versions of an image in the same fold.
Requested-part and alternate-part tokens require region occupancy of at
least 0.25; mixed tokens are excluded. The probe achieves a mean AUROC of 0.7962 across queries and request conditions, with balanced accuracy of 0.6996.

For each region, we compare its probe-predicted target probability when
requested and unrequested. We first average repeated part-pair comparisons
within each source query image, then bootstrap source query images.
The mean increase is 0.1841 (95\% CI: $[0.1677,0.2014]$).
The increase is positive for 94.0\% of source query images.

The expansion subset contains 83 pair-linked request comparisons,
corresponding to 81 unique task--query requests from 73 unique source
query images. Two task--query requests each participate in two different
part pairs. We average repeated comparisons within each source query image
before computing the mean change, the positive-change fraction, and the
bootstrap confidence interval, using these 73 images as the statistical
units. The mean increase is 0.3017 (95\% CI: $[0.2707,0.3318]$),
with positive changes for 98.6\% of source query images.
These results show that information
about the demonstrated target role remains readable across queries,
including when the output expands to the parent.

\subsection{SCPS settings and additional evaluation}
\label{app:scps-evaluation}

\subsubsection{Tasks and development data}
\label{app:tasks}

Table~\ref{tab:app-tasks} lists the four LIP tasks, 21 PartImageNet-OOD
tasks, and 22 Pascal-Part-116 tasks. They contain 400, 1,915, and 998
task--query pairs, respectively, totaling 3,313 pairs from 2,336 source images.

Each task provides eight demonstrations with target and parent annotations.
We select the palette and calibration weights on PACO-LVIS development data.
For each evaluated task, target quality is estimated on three labeled
development splits from its benchmark, with ten images per split except
Bicycle Seat, which has seven. We evaluate the same test queries with each
split's estimates. Test source images are disjoint from development and
demonstration images.

\begin{table}[!htbp]
\centering
\small
\caption{Tasks used in the final benchmark. Names retain the identifiers
in the evaluation records.}
\label{tab:app-tasks}
\begin{tabular}{@{}p{0.25\linewidth}p{0.70\linewidth}@{}}
\hline
Benchmark & Tasks \\
\hline
LIP & arms person; face person; legs person; shoes person \\
PartImageNet-OOD & Bicycle Seat; Biped Foot; Biped Hand; Biped Head;
Biped Tail; Bird Foot; Bird Head; Bird Tail; Bird Wing; Bottle Mouth;
Car Tier; Fish Fin; Fish Head; Fish Tail; Quadruped Foot; Quadruped Head;
Quadruped Tail; Reptile Foot; Reptile Head; Reptile Tail; Snake Head \\
Pascal-Part-116 & aeroplane:stern; aeroplane:wing; bird:head; bus:window;
car:window; cat:ear; cat:head; cat:leg; cat:neck; cat:paw; cat:tail;
cow:head; dog:ear; dog:head; dog:leg; dog:muzzle; horse:head; person:hair;
person:head; person:leg; person:lower arm; person:upper arm \\
\hline
\end{tabular}
\end{table}

\subsubsection{Palette and selection weights}
\label{app:palette-selection}

We select the parent-only color by mean STS across PACO-LVIS development
tasks, breaking ties by higher target IoU and then lower PL. The resulting
encoding uses white for the target, dark green $(0,0.16,0)$ for the
parent-only region, and black for the scene. We select $\alpha$ from
$\{0,0.25,0.5,1,2,4\}$ by development target IoU for each retrieval
method. Selected values are 4 for Appearance, CLIP, DINOv2, and UnsupPR;
0.5 for SupPR; and 1 for LaPR and RH-Partial2Global. All settings remain
fixed during test evaluation.

Both encodings of a demonstration share its normalized retrieval score but
have separate target quality estimates. For a fixed task and demonstration,
these estimates determine the preferred encoding. The retrieval score and
quality estimate together determine which demonstration is selected for a
query. The score includes no PL penalty.

Without target quality calibration, the binary and semantic contrast
versions of a demonstration receive the same retrieval score. Consistently
resolving these ties in favor of binary or semantic contrast therefore
recovers the corresponding uncalibrated single-encoding baseline in
Table~\ref{tab:scps_ablation} of the main paper.

\subsubsection{Output decoding}
\label{app:output-decoding}

We clip predictions to $[0,1]$ and assign pixels to color centers by squared
Euclidean RGB distance. Binary decoding uses white and black, assigning
exact ties to background. Semantic contrast decoding uses white, dark green
$(0,0.16,0)$, and black, with exact ties resolved in that order. Only pixels
assigned to white form the target mask. We apply no spatial postprocessing.

\subsubsection{Using the same decoder for both encodings}
\label{app:shared-decoder}

We decode the same RGB predictions with either white and black or white,
dark green, and black. The model, demonstrations, quality estimates,
selection weights, and selected pairs remain fixed. This control covers all
3,313 task--query pairs, seven retrieval methods, and three development
splits, using floating-point predictions before image quantization.

Table~\ref{tab:app-shared-decoder} shows that gains persist under either
shared decoder. Under white/black decoding, SCPS increases target IoU by
0.0544 with interval $[0.0393,0.0703]$, reduces PL by 0.0341 with
interval $[0.0134,0.0572]$, and increases STS by 6.42 percentage points
with interval $[4.54,8.41]$.

Changing only the decoder changes target membership for 0.154\% of
annotated target pixels and 0.062\% of parent-only pixels under original
selection, averaged equally across tasks. The corresponding SCPS shares
are 0.187\% and 0.061\%. The gains are therefore not explained by adding
the green decoding center.

\begin{table}[!htbp]
\centering
\caption{Original selection and SCPS under native or shared decoders.
Native decoding uses two centers for binary prompts and three for semantic
contrast prompts. All metrics are fractions; tasks are weighted equally.}
\label{tab:app-shared-decoder}
\begin{tabular}{llrrr}
\hline
Decoder & Selection & Target IoU & PL & STS \\
\hline
Native & Original & .2412 & .1512 & .1762 \\
Native & SCPS     & .2952 & .1168 & .2399 \\
White / black & Original & .2412 & .1512 & .1762 \\
White / black & SCPS     & .2956 & .1170 & .2404 \\
White / green / black & Original & .2407 & .1505 & .1761 \\
White / green / black & SCPS     & .2949 & .1164 & .2400 \\
\hline
\end{tabular}
\end{table}

\subsubsection{Where the separate color is assigned}
\label{app:color-location}

Table~\ref{tab:app-color-location} compares coloring the demonstrated
parent-only region with coloring a translated region of equal area outside
the parent. We start from the 200 intervention cases in
Appendix~\ref{app:feature-interventions}. Six do not permit an exactly
area-matched translated region and are excluded, leaving 194 cases across
37 tasks. The query, demonstration image, target, and palette stay fixed.
This comparison changes the demonstration encoding without running SCPS
selection.

Coloring the parent-only region reduces PL more than coloring the translated
region, with paired difference $-0.3095$ and bootstrap interval over tasks
$[-0.3671,-0.2519]$. It also reduces target recall. The effect depends on
where the color is assigned but can extend to the requested target,
supporting selection that considers target quality.

\begin{table}[!htbp]
\centering
\caption{Location of the colored region, averaged equally across 37 tasks comprising 194 cases with area-matched control regions. All metrics are fractions.}
\label{tab:app-color-location}
\begin{tabular}{lrrrr}
\hline
Encoding & Target IoU & Precision & Recall & PL \\
\hline
Binary                     & .2140 & .2261 & .8562 & .6681 \\
Correct parent-only region & .1896 & .2853 & .4728 & .3061 \\
Translated control region  & .2085 & .2266 & .7827 & .6157 \\
\hline
\end{tabular}
\end{table}

\subsubsection{Comparison with calibrated binary selection}
\label{app:calibrated-binary}

We compare SCPS with binary selection using the same target quality
calibration. After averaging queries and development splits within each
task, we compute paired differences and resample the 47 tasks.
Table~\ref{tab:app-calibrated-binary} reports gains from adding semantic
contrast candidates. Averaged across methods, target IoU increases by
0.0078, PL decreases by 0.0283, and STS increases by 1.56 percentage points.
All three mean intervals exclude zero; individual target IoU intervals for
UnsupPR, SupPR, and LaPR include zero.

\begin{table}[!htbp]
\centering
\small
\setlength{\tabcolsep}{4pt}
\caption{SCPS minus calibrated binary selection. Each entry gives a mean
paired task difference and its 95\% bootstrap interval over tasks.
IoU and PL differences are absolute fractions; STS differences are
percentage points. The mean row averages methods within each task before
resampling tasks.}
\label{tab:app-calibrated-binary}
\begin{tabular}{lccc}
\hline
Retrieval method & $\Delta$ Target IoU & $\Delta$ PL & $\Delta$ STS (pp) \\
\hline
Appearance & $+0.0104$ & $-0.0289$ & $+1.69$ \\
 & $[0.0032,0.0185]$ & $[-0.0461,-0.0142]$ & $[0.49,3.18]$ \\
CLIP & $+0.0094$ & $-0.0321$ & $+1.89$ \\
 & $[0.0012,0.0175]$ & $[-0.0515,-0.0148]$ & $[0.90,3.06]$ \\
DINOv2 & $+0.0086$ & $-0.0249$ & $+1.50$ \\
 & $[0.0024,0.0154]$ & $[-0.0400,-0.0115]$ & $[0.67,2.42]$ \\
UnsupPR & $+0.0061$ & $-0.0345$ & $+1.28$ \\
 & $[-0.0016,0.0145]$ & $[-0.0534,-0.0178]$ & $[0.32,2.45]$ \\
SupPR & $+0.0053$ & $-0.0244$ & $+1.61$ \\
 & $[-0.0018,0.0126]$ & $[-0.0394,-0.0115]$ & $[0.67,2.71]$ \\
LaPR & $+0.0053$ & $-0.0260$ & $+1.28$ \\
 & $[-0.0005,0.0117]$ & $[-0.0398,-0.0139]$ & $[0.52,2.15]$ \\
RH-Partial2Global & $+0.0092$ & $-0.0273$ & $+1.70$ \\
 & $[0.0012,0.0177]$ & $[-0.0485,-0.0079]$ & $[0.73,2.97]$ \\
\hline
Mean & $+0.0078$ & $-0.0283$ & $+1.56$ \\
 & $[0.0015,0.0146]$ & $[-0.0436,-0.0146]$ & $[0.74,2.58]$ \\
\hline
\end{tabular}
\end{table}

\subsubsection{Results for each benchmark and retrieval method}
\label{app:benchmark-method}

Table~\ref{tab:app-benchmark-method} expands the main benchmark averages.
Target IoU and STS improve for every listed method on each benchmark.
PL decreases for most combinations, with small increases in several settings
where baseline leakage is already low. Results average development splits
and weight tasks equally within each benchmark.

\begin{table}[!htbp]
\centering
\small
\setlength{\tabcolsep}{4pt}
\caption{Results for each benchmark for every retrieval method, averaged
equally across tasks within each benchmark. IoU denotes target IoU.
All metrics are fractions.}
\label{tab:app-benchmark-method}
\begin{tabular}{lrrrrrr}
\hline
& \multicolumn{3}{c}{Original binary} & \multicolumn{3}{c}{SCPS} \\
\cline{2-4}\cline{5-7}
Method & IoU & PL & STS & IoU & PL & STS \\
\hline
\multicolumn{7}{l}{\textbf{LIP}} \\
Appearance        & .3842 & .0152 & .4125 & .5476 & .0076 & .6292 \\
CLIP              & .3594 & .0160 & .3850 & .5321 & .0117 & .5892 \\
DINOv2            & .4072 & .0077 & .4500 & .5501 & .0091 & .6242 \\
UnsupPR           & .4165 & .0123 & .4575 & .5458 & .0071 & .6208 \\
SupPR             & .4090 & .0100 & .4550 & .4822 & .0094 & .5450 \\
LaPR              & .3883 & .0075 & .4350 & .4847 & .0094 & .5508 \\
RH-Partial2Global  & .4534 & .0102 & .4975 & .4965 & .0119 & .5567 \\
\hline
\multicolumn{7}{l}{\textbf{PartImageNet-OOD}} \\
Appearance        & .2117 & .2405 & .1259 & .2530 & .1845 & .1729 \\
CLIP              & .2160 & .2704 & .1195 & .2522 & .1715 & .1748 \\
DINOv2            & .2232 & .2560 & .1295 & .2574 & .1743 & .1744 \\
UnsupPR           & .2295 & .2441 & .1382 & .2588 & .1707 & .1794 \\
SupPR             & .2176 & .2549 & .1285 & .2240 & .2042 & .1467 \\
LaPR              & .2121 & .2427 & .1220 & .2329 & .2018 & .1471 \\
RH-Partial2Global  & .2501 & .2411 & .1547 & .2512 & .1709 & .1761 \\
\hline
\multicolumn{7}{l}{\textbf{Pascal-Part-116}} \\
Appearance        & .1838 & .0929 & .1148 & .3078 & .0688 & .2475 \\
CLIP              & .2265 & .0799 & .1711 & .3161 & .0713 & .2631 \\
DINOv2            & .2556 & .0773 & .1936 & .3251 & .0694 & .2710 \\
UnsupPR           & .2301 & .0821 & .1734 & .3168 & .0692 & .2627 \\
SupPR             & .2112 & .0779 & .1492 & .2518 & .0749 & .1998 \\
LaPR              & .2172 & .0785 & .1681 & .2759 & .0729 & .2300 \\
RH-Partial2Global  & .2815 & .0873 & .2263 & .3093 & .0879 & .2475 \\
\hline
\end{tabular}
\end{table}

\subsubsection{Comparison with uniform background recoloring}
\label{app:uniform-background}

We compare semantic contrast with a uniform-background encoding that assigns
white to the target and dark green $(0,0.16,0)$ to every non-target
demonstration pixel. Both policies retain binary candidates and use the same
demonstration pools, query-dependent selector scores, development splits,
and calibration procedure. The uniform-background outputs are decoded using
white, dark green, and black; using only white and dark green produces
identical aggregate metrics.

Table~\ref{tab:app-uniform-background} compares the candidate encoding sets.
Relative to uniform recoloring, semantic contrast changes PL by
$-0.0291$ (95\% CI: $[-0.0432, -0.0160]$), STS by
$+0.97$ percentage points (95\% CI: $[0.21, 1.83]$), and target IoU
by $-0.0020$ (95\% CI: $[-0.0071, 0.0031]$). Explicitly distinguishing the parent-only region
therefore improves average leakage control and strict target success beyond
uniform background recoloring.

\begin{table}[!htbp]
\centering
\small
\setlength{\tabcolsep}{4pt}
\caption{Comparison of candidate encoding sets with target quality
calibration. Results weight tasks equally and average across seven retrieval
methods. All metrics are fractions.}
\label{tab:app-uniform-background}
\begin{tabular}{@{}p{0.48\linewidth}rrr@{}}
\hline
Candidate encodings & Target IoU $\uparrow$ & PL $\downarrow$ & STS $\uparrow$ \\
\hline
Binary & .2875 & .1451 & .2243 \\
Binary + uniform background & .2973 & .1458 & .2303 \\
Binary + semantic contrast & .2952 & \textbf{.1168} & \textbf{.2399} \\
Binary + uniform background + semantic contrast & \textbf{.2976} & .1223 & .2398 \\
\hline
\end{tabular}
\end{table}

\subsection{Semantic distinctions in SINE}
\label{app:sine}

\begin{table}[!htbp]
\centering
\small
\caption{Dog versus scene AUROC in SINE under person requests, averaged
across 48 held-out queries. The feature row uses a linear probe; output
rows use the continuous score or binary mask directly.}
\label{tab:app-sine}
\begin{tabular}{@{}p{0.68\linewidth}r@{}}
\hline
Features or output & Mean query AUROC \\
\hline
Features after interaction with the demonstration & .988 \\
Continuous output score                           & .874 \\
Thresholded binary mask                           & .521 \\
\hline
\end{tabular}
\end{table}

SINE provides a complementary setting with native segmentation masks. With
demonstrations requesting the person, we train a linear probe to distinguish
dog from scene features on 80 queries and evaluate it on 48 disjoint queries.
Each evaluation query uses three demonstrations, whose results are averaged
within the query. We also measure this distinction using the continuous
output score and thresholded mask.

Table~\ref{tab:app-sine} shows that the dog remains distinguishable in
internal features, while the binary output largely merges it with the
scene. Both regions are background under a person request, so this is
consistent with the requested segmentation. Internal distinctions can
therefore persist even when output labels do not expose them. SINE's
restricted output encodings do not support the color comparisons we conduct with SegGPT.

\end{document}